\documentclass{article}
\usepackage{ijcai23}

\usepackage{times}
\usepackage{soul}
\usepackage{url}
\usepackage[hidelinks]{hyperref}
\usepackage[utf8]{inputenc}
\usepackage[small]{caption}
\usepackage{graphicx}
\usepackage{amsmath}
\usepackage{amsthm}
\usepackage{booktabs}
\usepackage{algorithm}
\usepackage{algorithmic}
\usepackage[switch]{lineno}
\usepackage{subcaption}
\usepackage{color, colortbl}
\usepackage{booktabs, multirow}
\usepackage{amssymb}
\usepackage{pifont}

\newcommand{\etal}{\textit{et al.}}
\newcommand{\xmark}{\ding{55}}
\newcommand{\dtoprule}{\specialrule{1pt}{0pt}{0.4pt}%
            \specialrule{0.3pt}{0pt}{\belowrulesep}%
            }
\newcommand{\dbottomrule}{\specialrule{0.3pt}{0pt}{0.4pt}%
            \specialrule{1pt}{0pt}{\belowrulesep}%
            }
\title{A HRNet-based Rehabilitation Monitoring System}

\author{
Yi-Ching Hung
\and
Yu-Qing Jiang
\and
Fong-Syuan Liou
\and
Yu-Hsuan Tsao
\and
\\Zi-Cing Chiang 
\And
Min-Te Sun
\affiliations
Department of Computer Science and Information Engineering\\ National Central University, Taiwan
\emails
\{jenny120244, e890316, jenny20030314, selina920120\}@gmail.com,\\ a108502552@g.ncu.edu.tw, msun@csie.ncu.edu.tw
}

\begin{document}

\maketitle

\begin{abstract}
    Rehabilitation treatment helps to heal minor sports and occupational injuries. In a traditional rehabilitation process, a therapist will assign certain actions to a patient to perform in between hospital visits and will rely on the patient to remember actions correctly and the schedule to perform them. Unfortunately, many patients forget to perform actions or fail to recall actions in detail. As a consequence, the rehabilitation treatment is hampered or, in the worst case, the patient may suffer from additional injury caused by performing incorrect actions. To resolve these issues, we propose a HRNet-based rehabilitation monitoring system, which can remind a patient when to perform the actions and display the actions for the patient to follow via the patient's smartphone. In addition, the system helps the therapist to monitor the progress of rehabilitation for the patient. Our system consists of an iOS app and several components at the server side. The app is in charge of displaying and collecting action videos. The server computes the similarity score between the therapist's actions and the patient's in the videos to keep track of the number of repetitions of each action. These stats will be shown to both the patient and therapist. The extensive experiments show that the F1-Score of the similarity calculation is as high as 0.9 and the soft accuracy of the number of repetitions is higher than 90\%.
\end{abstract}

\section{Introduction}
\label{sec:Introduction}
Nowadays, more and more people learn to utilize rehabilitation treatment to recover from minor occupational and sports injuries and improve the quality of their lives. In a rehabilitation process, a patient will work closely with her therapist and visit the therapist on a weekly basis to check the progress. During the visit, the therapist will instruct the patient to perform certain actions a number of times for rehabilitation at home. Traditionally, therapy relies on the patient to remember the actions and perform them correctly in between visits. There are two possible issues in this approach. First, if a patient forgets the actions or does not perform the actions for the treatment, it may delay the progress of rehabilitation process. Second, a patient who performs incorrect actions without realizing it may cause more damage. Therefore, there is a strong demand for a monitoring system between the therapist and the patient, which can remind the patient to perform correct actions and help the therapist to monitor how closely the patient follows the instructions.

In summary, the contributions of this research are listed as follows.
\begin{itemize}
    \item We propose an HRNet-based monitoring system that can be used in rehabilitation treatment between the therapist and the patient.
    \item We identify incorrect coordinates from HRNet~\cite{Sun2019HRNet} pose detection by using outlier detection and the exploitation of temporal correlation and fix the incorrect coordinates by extrapolation.
    \item We apply KL-divergence to calculate the similarity score between the poses in two different videos.
    \item We utilize the S-G filter~\cite{savitzky64} and Python's peak function to calculate the number of repetitions for an action in a video.
    \item We conduct extensive experiments on the proposed system. The results indicate that the F1-Score of the similarity calculation is above 0.9 and the soft accuracy of the number of repetitions is above 90\%.
\end{itemize}

The rest of this paper is organized as follows. In Section~\ref{sec:RelatedWork}, the prior works on pose detection and its applications are reviewed. In Section~\ref{sec:Design}, the design of our HRNet-based rehabilitation monitoring system is elaborated. Section~\ref{sec:Performance} reports the performance of the proposed system. Finally, Section~\ref{sec:Conclusion} provides the conclusion of this research.

\section{Related Work}
\label{sec:RelatedWork}
\subsection{2D Pose Detection}
Traditional approaches to articulated pose estimation have often used pictorial structure models~\cite{Felzenszwalb03pictorialstructures,Ramanan2005Strikeapose,Andriluka2009Pictorialstructures,Pishchulin2013Poselet}, in which spatial relationship between the parts in the human kinematic chain are represented as a tree-structured graphical model.
% deep pose 2014
In order to boost the accuracy of pose recognition, most approaches employ Convolutional Neural Network (CNN) architectures. For instance, DeepPose~\cite{Toshev2014DeepPose} is one of the fundamental methods for human pose estimation. In~\cite{Toshev2014DeepPose}, the authors formulate the pose estimation as a CNN-based regression problem to joint coordinates.
% alpha pose 2018
For multi-person pose estimation, ~\cite{Fang2017alphapose} further use top-down methods that first employ a person detector, followed by applying a pose estimator to calculate the pose for each person. However, the top-down methods are sensitive to the shift of the bounding box and cannot handle the case of occlusion and overlapping. Moreover, the runtime of their system is very expensive since it requires additional effort to separate each person in the image before determining the pose of each individual.
% open pose 2019
By contrast, bottom-up methods first detect all body joints in the image, then group them into single-person. One of the most relevant works is introduced by~\cite{Cao2019OpenPose}. They define a real-time 2D human pose system, known as OpenPose. The model uses a non-parametric representation, named Part Affinity Fields (PAFs), to predict vector fields and estimate association with individuals in the multi-person images.
% blaze pose 2020
In~\cite{bazarevsky2020blazepose}, the authors introduce a lightweight CNN architecture for on-device, real-time, single person-specific body pose tracking system, BlazePose. Their pose estimation uses heatmaps to supervise the lightweight embedding and regression to joint coordinates.
In~\cite{Sun2019HRNet}, a novel pose detection architecture, named HRNet, is introduced, which focuses on maintaining reliable high-resolution representations and repeatedly conducting multi-scale fusions across parallel convolutions for spatially precise heatmap estimation. The network contains both high resolution subnetworks and compressed resolution subnetworks. To aggregate the information from high to low resolutions of subnetworks in HRNet, an exchange unit is introduced.

\subsection{Applications of Human Pose Estimation}
Human pose estimation has applications in various fields, such as rehabilitation system, yoga, sports, and fitness, which are presented below.
% rehabilitation system
Li \etal~\cite{Li2020inhome} develop an in-home human pose estimation based lower body rehabilitation system. In~\cite{Li2020inhome}, they design a light-weight CNN model in order to keep a mobile device running smoothly. The challenge with this method is insufficient rehabilitation detection dataset and 3D location information to improve deep CNN model for their device.
% Yoga, sport and fitness
Different from~\cite{Li2020inhome}, some previous works~\cite{Huang2020Yoga,Rishan2020Yoga} consider the issue whether a trainee is able to follow the correct yoga or fitness posture. Hence, they introduce yoga assistant system to analyze and evaluate the postures of trainees in order to avoid injuries. Another relevant work, such as the AI Coach system~\cite{Wang2019AICoach}, provides personalized athletic training experiences by using spatial-temporal relation module to detect human pose. In~\cite{Khurana2018GymCam}, the authors present a system that tracks multiple human motions in a gym concurrently. Similar to~\cite{Khurana2018GymCam},~\cite{alatiah2020recognizing} also use pose estimation to recognize physical exercises and detect the number of repetitions.

\section{Design}
\label{sec:Design}
\begin{figure}
    \begin{center} 
        \includegraphics[width=3in]{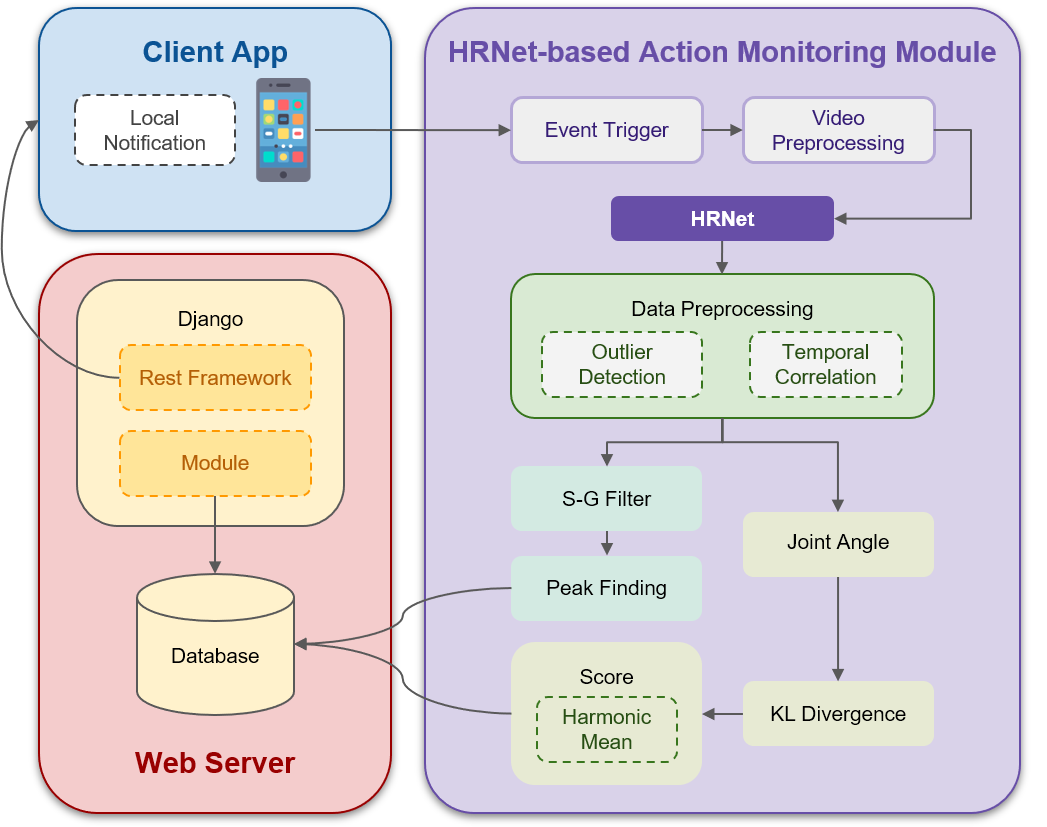}
        \caption{Architecture of the HRNet-based rehabilitation monitoring system.} \label{mysystem}
    \end{center}
\end{figure}

In this section, we introduce our HRNet-based rehabilitation monitoring system. The architecture of our system is shown in Figure~\ref{mysystem}. As can be seen in the figure, our system can be divided into three blocks. The blue block represents the client app, the pink block shows the web server, and the purple block is the HRNet-based action monitoring module. 
In the following sections, each block will be further elaborated in detail.

\subsection{Client App}
\begin{figure}
    \begin{center} 
        \includegraphics[width=2.1in]{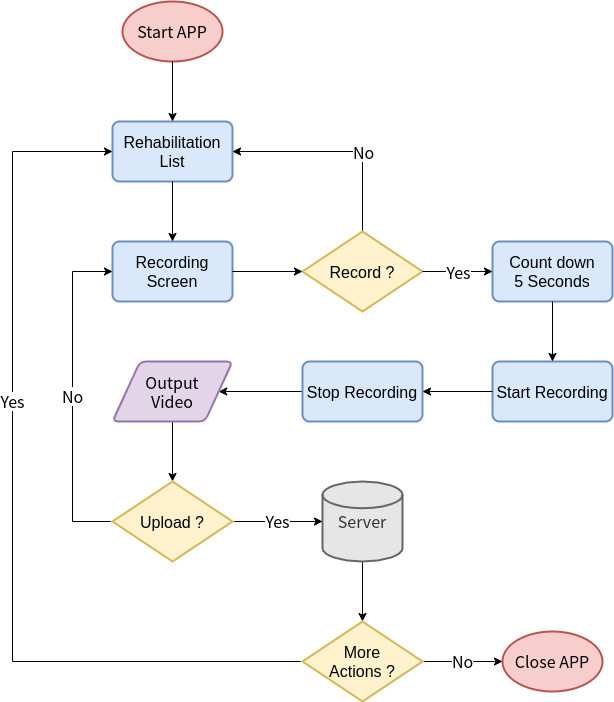}
        \caption{Flow diagram of our app process.} \label{app_process}
    \end{center}
\end{figure}

We develop a client app in Swift, which is recommended for native mobile application development on iOS by Apple. In Figure~\ref{app_process}, the operating process of our app is introduced.

\begin{figure*}[h]
     \centering
     \begin{subfigure}[t]{0.155\textwidth}
         \includegraphics[width=\textwidth]{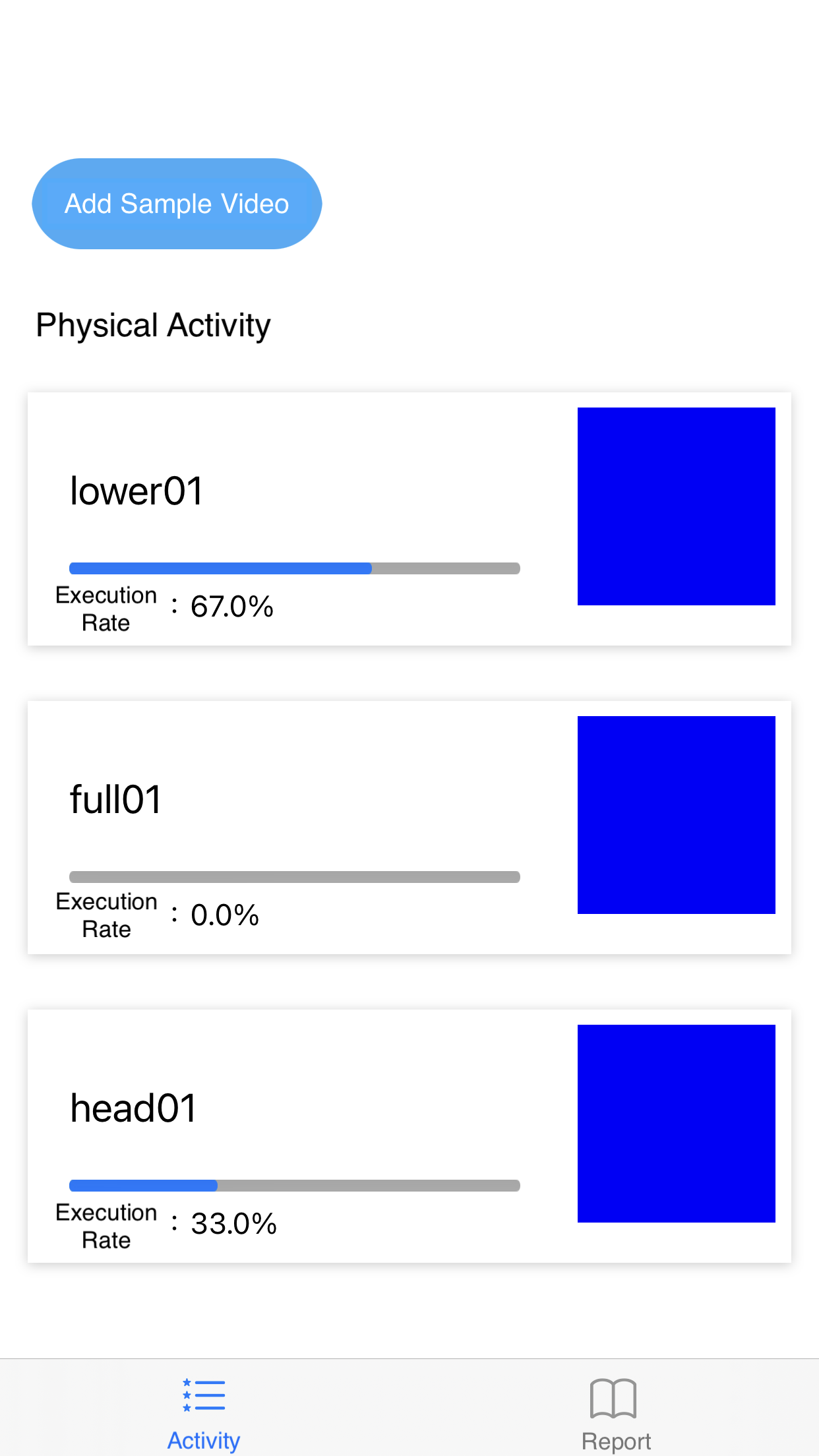}
         \caption{Rehabilitation actions.}
         \label{fig:r_list}
     \end{subfigure}
     \hspace{0cm}
     \begin{subfigure}[t]{0.155\textwidth}
         \includegraphics[width=\textwidth]{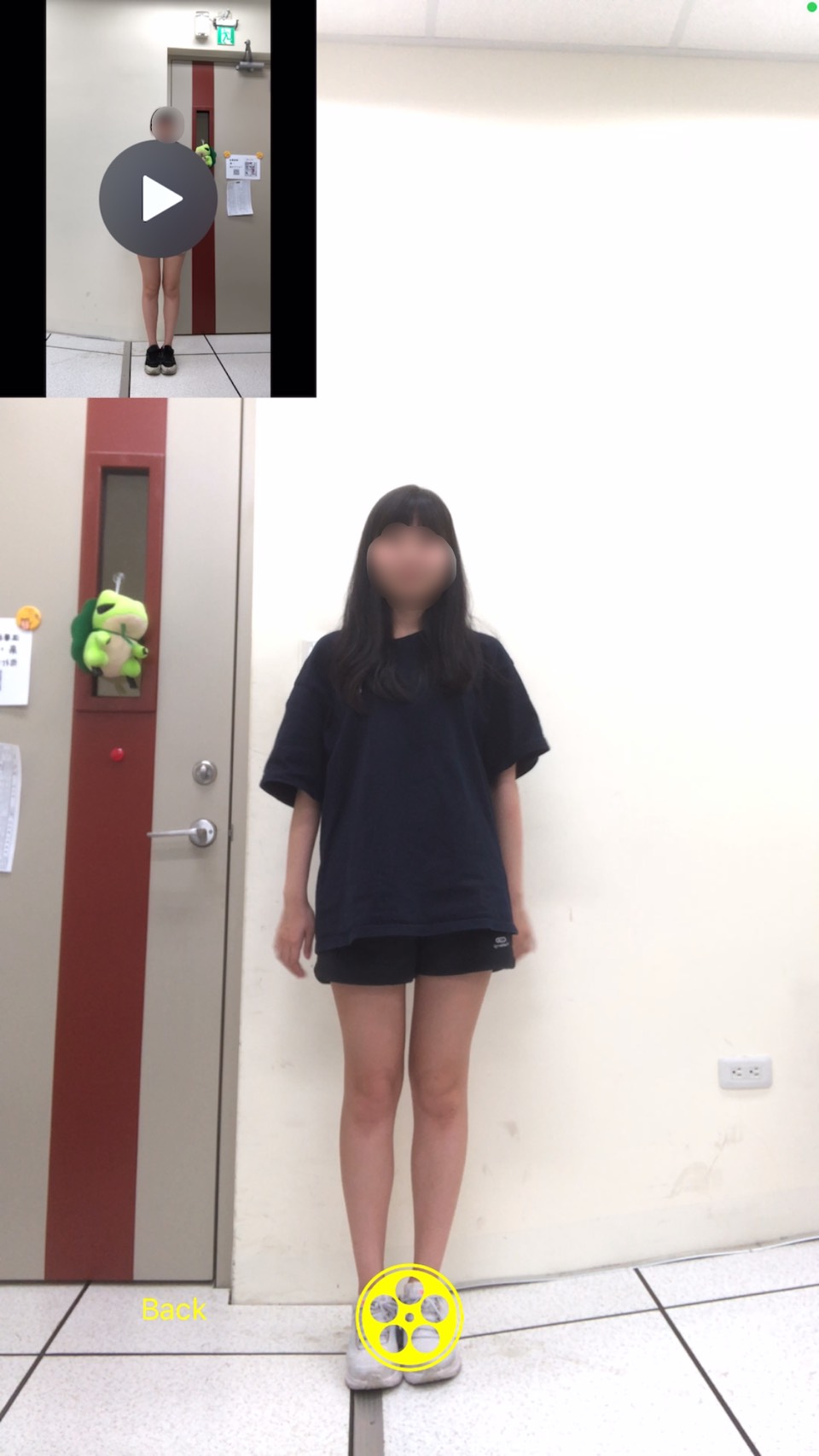}
         \caption{Record page.}
         \label{fig:record_screen}
     \end{subfigure}
     \hspace{0cm}
     \begin{subfigure}[t]{0.155\textwidth}
         \includegraphics[width=\textwidth]{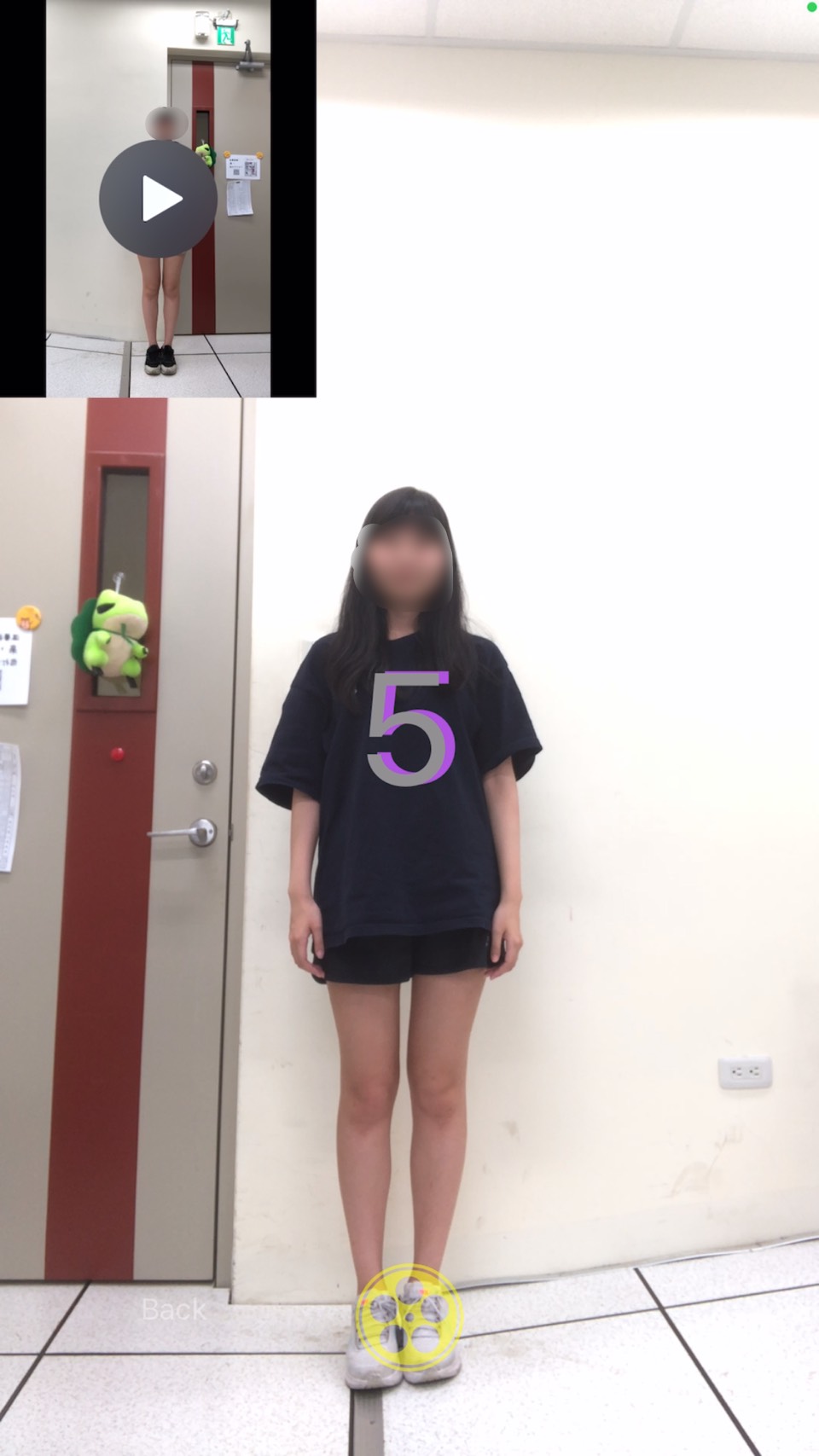}
         \caption{Countdown timer.}
         \label{fig:count_screen}
     \end{subfigure}
     \hspace{0cm}
     \begin{subfigure}[t]{0.155\textwidth}
         \includegraphics[width=\textwidth]{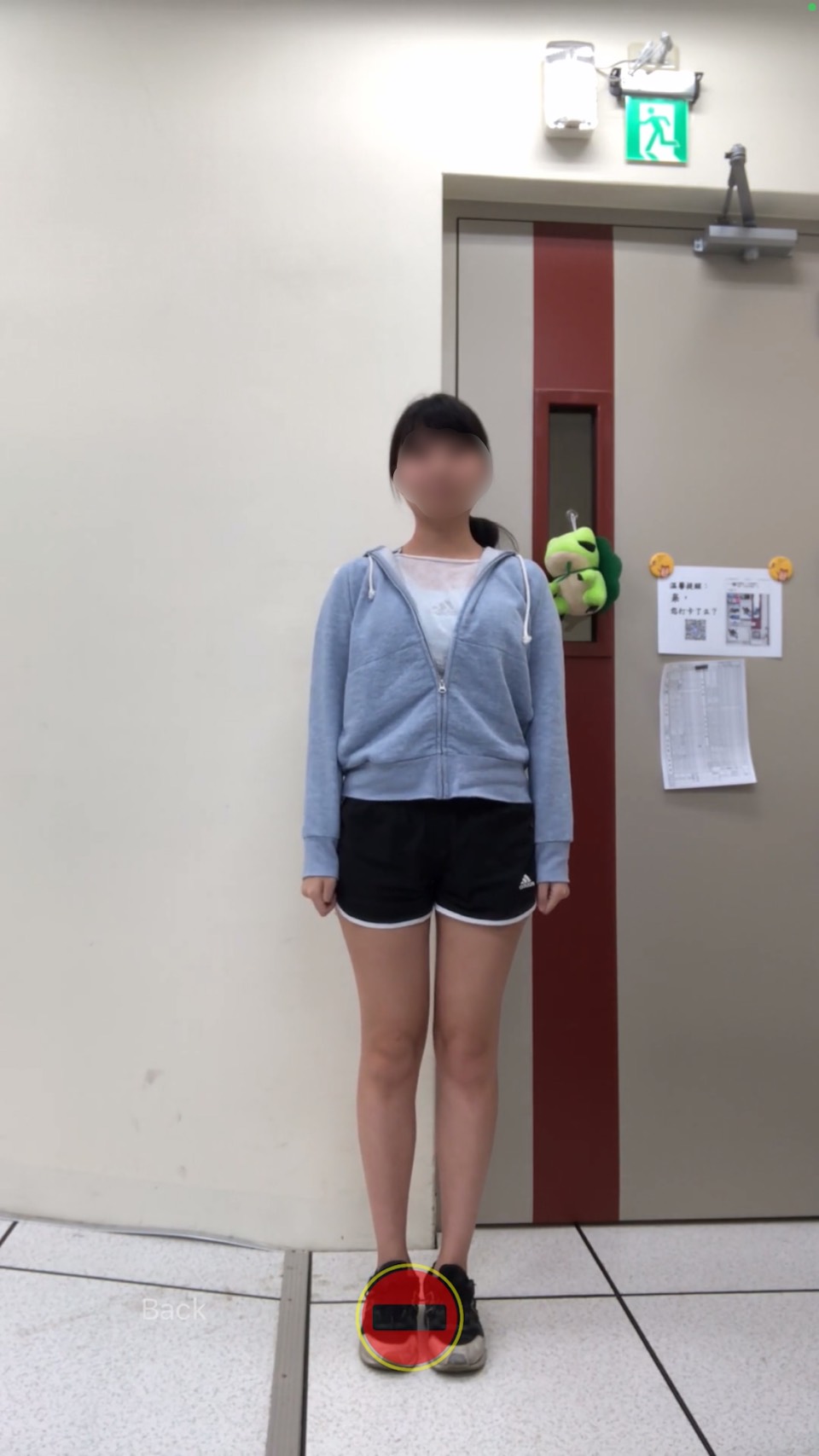}
         \caption{Start recording.}
         \label{fig:start_record}
     \end{subfigure}
     \hspace{0cm}
     \begin{subfigure}[t]{0.155\textwidth}
         \includegraphics[width=\textwidth]{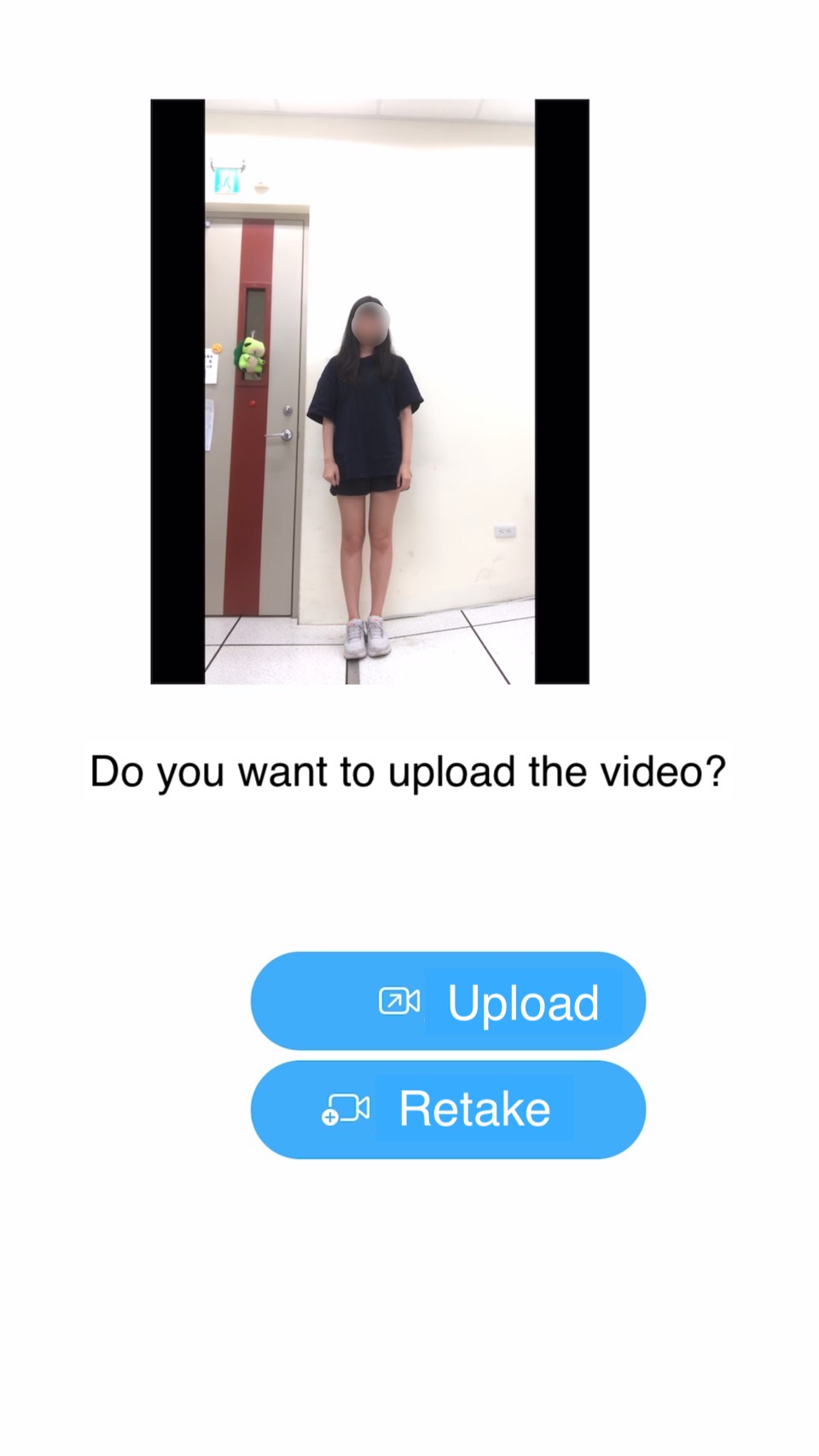}
         \caption{Upload page.}
         \label{fig:upload_screen}
     \end{subfigure}
     \hspace{0cm}
     \begin{subfigure}[t]{0.155\textwidth}
         \includegraphics[width=\textwidth]{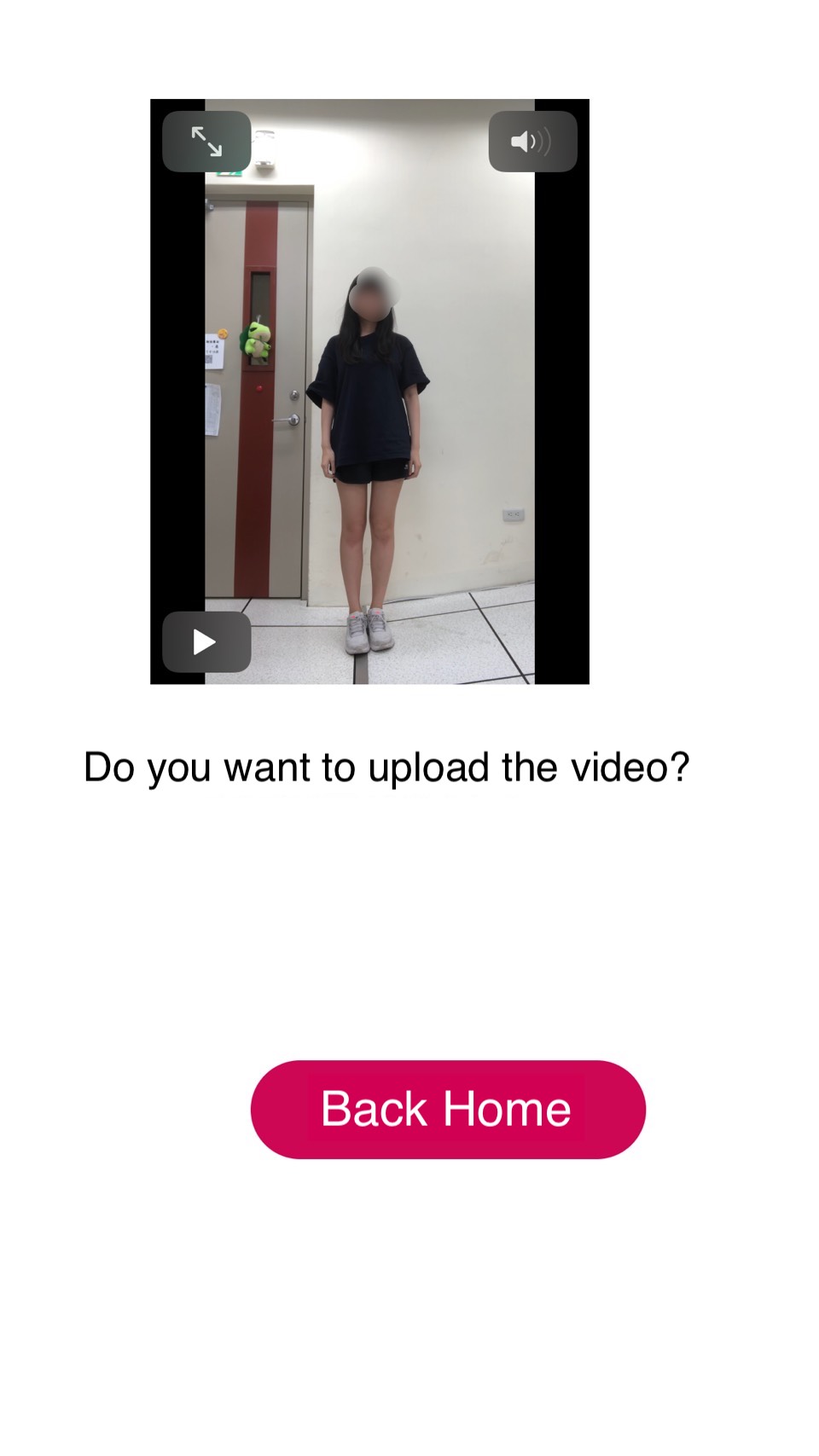}
         \caption{Finish uploading.}
         \label{fig:finish_upload}
     \end{subfigure}
     \caption{App user interface.} \label{fig:App_screen}
\end{figure*}

First of all, all the rehabilitation actions that need to be done are listed on the smartphone screen, as shown in Figure~\ref{fig:r_list}. The patient can choose one of the actions to go to the record page, as shown in Figure~\ref{fig:record_screen}. As can be seen in the figure, the sample video is at the top left of the screen and the record button is at the bottom. The patient can watch the sample video before recording his/her action. After pressing the record button, the screen will display a countdown timer, as shown in Figure~\ref{fig:count_screen}.

The app plays the sample video and records the patient’s action at the same time, as shown in Figure~\ref{fig:start_record}. After the camera stops recording, the app navigates to the upload page, which contains the upload button and the \emph{record again} button, as shown in Figure~\ref{fig:upload_screen}. The patient can watch and confirm the recorded video and press either the upload button or the \emph{record again} button. If the patient presses the upload button, the recorded video will be uploaded to our server, and the app will go back to the list of actions, as shown in Figure~\ref{fig:finish_upload}.

\subsection{Web Server}
Web server is used to store videos and provide functions. There are five modules developed by Django. Each of them will be described as follows.

    \definecolor{check}{rgb}{0.85,0.917,0.827}
    \begin{table}[!htb]
    \centering
    \caption{An example of a rehabilitation action.}
        \begin{tabular}{c c c c c c c}
            \hline
            \dtoprule
            \textbf{Day} & \textbf{1} & \textbf{2} & \textbf{3} & \textbf{4} & \textbf{5} & \textbf{6}\\
            \hline
            \multirow{4}{*}{\textbf{Repetition}} 
            &\multicolumn{1}{c}{13} & \multicolumn{1}{c}{10} &
            \multicolumn{1}{c}{5} & \multicolumn{1}{c}{8} & \multicolumn{1}{c}{0} & \multicolumn{1}{c}{5}\\
            
            & \multicolumn{1}{c}{10} & \multicolumn{1}{c}{11} & \multicolumn{1}{c}{12} & \multicolumn{1}{c}{5} & \multicolumn{1}{c}{0} & \multicolumn{1}{c}{10}\\
            
            & \multicolumn{1}{c}{8} & \multicolumn{1}{c}{10} & \multicolumn{1}{c}{10} & \multicolumn{1}{c}{0} & \multicolumn{1}{c}{0} & \multicolumn{1}{c}{10}\\
            
            & \multicolumn{1}{c}{5} & \multicolumn{1}{c}{0} & \multicolumn{1}{c}{0} & \multicolumn{1}{c}{0} & \multicolumn{1}{c}{0} & \multicolumn{1}{c}{10}\\
            \hline
            \textbf{Pass?} & \xmark & \cellcolor{check}\large\checkmark & \xmark & \xmark  & \xmark  & \cellcolor{check}\large\checkmark\\
            \hline
            \multicolumn{1}{c}{\textbf{Completion Rate}} & \multicolumn{6}{c}{66.7\%}\\
            \hline
            \dbottomrule
        \end{tabular}
    \label{tab:action_ex}
    \end{table}

\begin{itemize}
    \item \textit{Rehabilitation action module:} The functions of this module include adding a rehabilitation action, notifying the patient of rehabilitation, and calculating the completion rate. The daily checkpoint is defined as a fixed number of sets of an action a patient needs to complete in a day. Each action can be assigned by the therapist to be light, medium, or daily. If an action is assigned to be light, the patient is required to collect 3 daily checkpoints each week. If an action is assigned to be medium, the patient is required to collect 5 daily checkpoints each week. If an action is assigned to be daily, the patient is required to collect 7 daily checkpoints each week. The completion rate is then defined as the number of collected daily checkpoints divided by the number of required checkpoints each week. 
    
    The patient notification is set based on the completion rate, which will be executed every 24 hours after the app is started the first time (a timer will be running continuously at the background mode). At the beginning, the patient ID is obtained from the app. The patient ID is used to query patient's rehabilitation action list, which includes the detailed information of each action, such as start date, the daily checkpoint value, and how many daily checkpoints to be collected per week. The daily checkpoint and the start date for each action are initialized by the therapist. The detection list of each action contains the detection results of all uploaded videos from the previous day, which include the total number of repetitions for the action of the previous day. If the total number of repetitions is higher than 10, it means that the patient completes one set of actions. Then, we can calculate the completion rate of the previous day. If today is not the day to visit the therapist and the completion rate of the previous day is less than 100\%, the notification will be triggered.
    
    An example of a rehabilitation action is shown in Table~\ref{tab:action_ex}. In this example,, the action is assumed to be assigned by the therapist to be light, which means that the patient is required to collect 3 daily checkpoints before she sees the therapist again. On Day 2, the patient submits three videos. The detection results of the three videos are 10, 11, and 10 number of repetitions of the action, respectively. Since there are 3 sets of action with higher than or equal to 10 repetitions, as a consequence, the patient earns a daily checkpoint on Day 2. On Day 6, the patient submits 4 videos. Although the first video has only 5 repetitions, the last three have enough repetitions for the patient to earn another daily checkpoint. Except for Day 2 and Day 6, the patient either submits fewer than 3 videos or does not have enough repetitions in 3 videos to earn any daily checkpoint. One interesting observation is that the patient actually has more than 30 total repetitions of the action on Day 1, but still fails to earn a daily checkpoint due to the specific requirement of daily checkpoint. In summary, the patient in this example collects a total of 2 daily checkpoints, and her completion rate is 66.7\%. Since the patient is yet to see the therapist again and has not had 100\% completion rate, she will receive a notification as a reminder for more rehabilitation exercise.

    \item \textit{User module:}
    The goal of this module is to add users and get the user information.
    \item \textit{Recorded video module:}
    All functions related to the recorded video are included in this module, such as uploading videos and getting a list of all videos uploaded by users.
    \item \textit{Sample video module:}
    This module is used to upload sample videos and get the sample video information. In addition, the most important function is to serve video streaming.
    \item \textit{Detection result module:}
    The detection results will be added to the database by this module. Therefore, we can use the function of this module to query the detection results of the user's videos.
\end{itemize}

\subsection{HRNet-Based Action Monitoring Module}
This module is the core of our HRNet-based rehabilitation monitoring system. This module is composed of video processing, data preprocessing, and similarity calculation. In video processing, HRNet is used to extract pose features from the patient's uploaded video. In data preprocessing, outlier detection and temporal correlation are used to detect and correct some pose features wrongfully detected by HRNet. Finally, in similarity calculation, the similarity and the number of repetitions of each action is calculated and recorded to monitor the patient's rehabilitation. These components are elaborated in detail as follows.

\subsubsection{Video processing}
An event trigger in the video upload folder is created so that new uploaded videos will trigger the following actions. 

First, since the video format is an open container format developed by Apple, it needs to be converted to MP4 format. Next, HRNet-W32 is used to perform human pose estimation in the video, where 32 represents the width of the high-resolution subnetworks in the last three stages of HRNet. HRNet detects all of the bounding boxes, each corresponding to a human, and the 17 Keypoints in each bounding box. Table~\ref{tab:keypoint} shows the joint nomenclature corresponding to the 17 keypoints.

\setlength{\tabcolsep}{8pt}
\begin{table}[!htb]
\centering
\caption{A depiction of the 17 joint keypoints.}
    \begin{tabular}{c c | c c}
        \hline
        \textbf{0} & Nose          &             & \\
        \textbf{1} & Left eye      & \textbf{2}  & Right eye \\
        \textbf{3} & Left ear      & \textbf{4}  & Right ear\\
        \textbf{5} & Left shoulder & \textbf{6}  & Right shoulder\\
        \textbf{7} & Left elbow    & \textbf{8}  & Right elbow\\
        \textbf{9} & Left wrist    & \textbf{10} & Right wrist\\
        \textbf{11} & Left hip     & \textbf{12} & Right hip\\
        \textbf{13} & Left knee    & \textbf{14} & Right knee\\
        \textbf{15} & Left ankle   & \textbf{16} & Right ankle\\
        \hline
    \end{tabular}
\label{tab:keypoint}
\end{table}

\subsubsection{Data Preprocessing} \label{sec:dp}
In the coordinate dataset, a fraction of the keypoint coordinates may be detected by HRNet incorrectly. To deal with this issue, the incorrect locations need to be identified and revised appropriately. Two different methods are used to identify the incorrect coordinates. The first one is outlier detection, and the second one makes use of temporal correlation.

First, the difference of the corresponding keypoint coordinates between the consecutive frames, i.e., keypoint displacement, is calculated. For the two consecutive frames for keypoint displacement computation, the later frame is referred to as the primary frame, and the earlier frame the secondary frame. Note that the displacement of a keypoint should be within a reasonable range. In other words, the outliers of the displacements are identified to be incorrect.

To remove outliers, Z-Score of each displacement is calculated. If the absolute value of the Z-Score value of a displacement is greater than 3, i.e., if the displacement is outside of the range between the average of displacements plus and minus three standard deviations, it is considered as an outlier. 
Given an outlier of a keypoint displacement, we first identify the corresponding primary frame, and replace the coordinates of the keypoint in the primary frame by the average of the coordinates of the same keypoint in the secondary frame and in the following frame of the primary frame.

After removing outliers, there may still be incorrect coordinates. In fact, from our experiments with HRNet, we found that it may output incorrect coordinates for a keypoint in a number of consecutive frames (up to 4 frames). To further identify these incorrect coordinates, the temporal correlation of consecutive frames is exploited. To achieve this goal, the maximum displacement of each keypoint in a video is first computed after the removal of outliers. To determine if a displacement is too large, we define a displacement threshold \(T\)\% for all keypoints. A displacement of a keypoint is considered too large if the displacement is larger than \(T\)\% of the maximum displacement of the keypoint. There are two possible  explanations for such a large displacement. Either the patient is performing a large range of motion, or one coordinate of the keypoint in previous or current frames is incorrect.

To identify which explanation is more plausible, the most recent six displacements of the same keypoint (i.e., 3 prior to the current frame and 3 after) will be examined. To be specific, when the displacement of a keypoint is large, the determination of incorrect coordinates for the keypoint is as follows. We first verify if the displacements of the keypoint in the previous three frames are small and the displacements of the keypoint in the next three frames are large. If this condition is satisfied, we consider that the patient is starting a large range of motion and conclude that the coordinates of the keypoint in the current frame are correct. If this is not the case, we verify if the displacements of the keypoint in the previous three frames are large and the displacements of the keypoint in the next three frames are small. If this condition is satisfied, we consider that the patient is finishing a large range of motion and conclude that the coordinates of the keypoint in the current frame are correct. If this is again not the case, we verify if the displacements of the keypoint in the previous three frames and the next three frames are all large. If this condition is satisfied, we consider that the patient is in the middle of performing a large range of motion and conclude that the coordinates of the keypoint in the current frame are correct. All other cases excluding the previous three satisfy at least one of the following three conditions: 1) the displacements of the keypoint in the previous three frames are mixed with large and small; 2) the displacements of the keypoint in the next three frames are mixed with large and small; 3) the displacements of the keypoint in the previous three frames and next three frames are all small. (Since the displacements of the keypoint in the current frame are large, there will still be a mixture of small and large displacements in consecutive frames.) These mixed small and large displacements in the consecutive frames indicate that the keypoint is suddenly moving at a high speed within a very short period of time, which is unlikely for the rehabilitation process. (The duration of two consecutive frames is only 0.1 second.) Hence, in the case when the small and large displacements of the keypoint are mixed in consecutive frames, we conclude that the coordinates of the keypoint in the current frame are incorrect.
When the coordinates of a keypoint in the current frame are identified as incorrect, it should be revised appropriately. Since HRNet may output incorrect coordinates of a keypoint in a number of consecutive frames, using interpolation for the revision is not a good idea. Instead, the coordinates of the keypoint in the previous two frames are extrapolated to estimate the coordinates of the keypoint in the current frame.

\subsubsection{Similarity Calculation}
In this section, we use the pre-processed keypoint coordinates to calculate the score of pose similarity and the number of repetitions for an action. The former is calculated by comparing the video uploaded by the patient with the sample video provided by the therapist. The latter focuses on the video which is uploaded by the patient. These tasks are elaborated in detail as follows.

\begin{itemize}
    \item \textit{Pose similarity:} \\
    Our goal is to calculate the the score of pose similarity for an action between the sample video provided by the therapist and the recorded video uploaded by the patient. To achieve this goal, the first is to obtain the angles of human joints. Figure~\ref{fig:angle_ex} illustrates the angles between multiple sets of vectors obtained by pre-processed keypoint coordinates.
    
        \begin{figure}[H]
            \begin{center} 
                \includegraphics[width=1.2in]{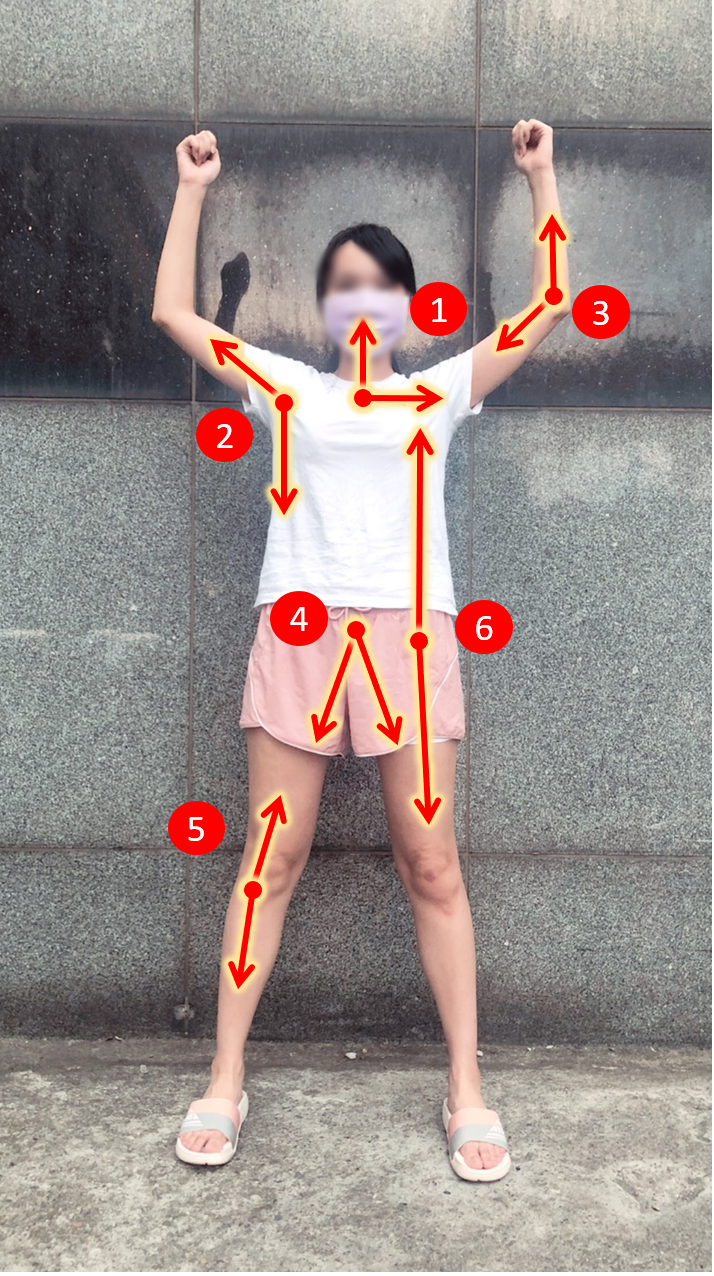}
                \caption{Angles between multiple sets of vectors.} \label{fig:angle_ex}
            \end{center}
        \end{figure}

    First, the arccos function is used to return the angle of two vectors $A$ and $B$, which is calculated by Equation~\ref{arccos}. 
        
        \begin{equation}
        \begin{aligned}
        \label{arccos} %arccos
            \theta
            & = arccos(\frac{ \sum\limits_{i=1}^{n}{{\bf A}_i{\bf B}_i} }{ \sqrt{\sum\limits_{i=1}^{n}{{\bf A}_i^2}} \sqrt{\sum\limits_{i=1}^{n}{{\bf B}_i^2}} }),
        \end{aligned}   
        \end{equation}

    where $A_i$ and $B_i$ are components of vector $A$ and $B$ respectively.

    In order to quantify the similarity between the patient’s pose and the pose in the sample video, KL divergence~\cite{Kullback1968} is used. KL divergence can measure the divergence of a specific distribution from another reference distribution. Given a specific angle, its variation over time in the patient's uploaded video is treated as a distribution and the variation of the same angle over time in the sample video is treated as the reference distribution. Then, the KL divergence of these two distributions is used to measure the divergence between them. In total, there are 11 KL divergences obtained from 11 different angles.
    
    In theory, if the KL divergence of an angle is close to 0, it implies that the patient's pose is similar to the sample video for this angle. On the contrary, if the KL divergence of an angle is large, it implies that the patient's action is different from the sample video for this angle. However, the since KL divergence is unbounded, it is impossible to find a middle value as the threshold to determine if patient's action is correct. In order to calculate the score of pose similarity using KL divergence values, the upper bound (i.e., maximum value) of KL divergence for each angle is obtained from the 50 KL divergences of the angle measured from 50 videos with incorrect poses. The upper bound provides us a reference KL value to the incorrect pose.
    
    After the upper bound of 11 angles is obtained, the score of each angle is calculated using Equation~\ref{my_score}. The range of the score for an angle is rescaled to be from 0 to 100, where 0 means the patient's action is very different from the sample video for this angle, and 100 means the patient's action is identical to the sample video for this angle.
    
        \begin{equation}
        \label{my_score}
            {\bf score} = max(100 - \frac{100 \times d}{u} , 0),
        \end{equation}
        
        where $d$ is the KL divergence of the angle and $u$ is the upper bound of the angle.
    
    Finally, the overall similarity score is calculated as the harmonic mean of the scores of all 11 angles. The reason for using harmonic mean instead of arithmetic average is because, in the case when most scores are high and others are low, it may indicate that this action is focusing on a certain part of the body (e.g., head or shoulder movement only). In this case, the few low scores may come from the patient's incorrect action on the focusing part of the body, and all the other high scores may come from the angles associated with the other parts of the body that are not moving at all. If the arithmetic mean is used, the resulting overall score will be relatively high because only a few scores are low. However, in a case like this, the overall score should be low because the patient is making incorrect movements at the focusing part of the body. As a consequence, the harmonic mean is used to compute the overall score, which can avoid the aforementioned issue.

    Given positive real numbers \(x_1, x_2, ..., x_n\), their harmonic mean \(H\) can be computed by Equation~\ref{harmonic}.
        \begin{equation}
        \label{harmonic}
            H = \frac{n}{\frac{1}{x_1} + \frac{1}{x_2} + ... + \frac{1}{x_n}}
        \end{equation}
        
    \item \textit{Number of repetitions:} \\
    In order to obtain the number of repetitions for an action in a patient's uploaded video, we first need to obtain the number of cycles for each keypoint. For a given keypoint, the first frame of the video is used as the reference frame, and then the displacements of the keypoint between the following frames and the reference frame are calculated. After that, we obtain a series of displacements for the keypoint. In general, finding the number of cycles in a series of displacements for a keypoint is equivalent to finding the number of local maximums (i.e., peaks) in the same series. To do that, the S-G filter~\cite{savitzky64} is used first to smooth the series so that the peaks are easier to find. Then, the python function \(scipy.signal.find_peaks_cwt()\) is called to identify the number of peaks in the series with wavelet transformation.
    
    After obtaining the number of cycles for each keypoint in a patient's uploaded video, we first compute a set of modes for all keypoints. Then, the number of repetitions for the action is the lowest number in the set. In theory, it seems to be more reasonable to use the greatest common divisor among the numbers of cycles for all keypoints as the number of repetitions. However, from extensive experiments, we found the two following reasons to adopt the lowest mode instead of the greatest common divisor:
    
        \begin{enumerate}
            \item The number of cycles for a keypoint may not be very accurate. When there exist errors among the numbers of cycles for all keypoints, the greatest common divisor will be very different from the true number of repetitions.
            \item There are totally 17 keypoints. With these many keypoints, the chance of some keypoint to have only one cycle in one repetition of action is very high. Hence, using the lowest mode is reasonable.
        \end{enumerate}
    
    \end{itemize}

\section{Performance}
\label{sec:Performance}
\subsection{Experimental Environment}
In our experiments, the hardware specification of our server is shown in Table~\ref{tab:server_spec}. We use Xcode on a MacBook Air to build the App for iOS.  Our App runs on an iPhone 8 Plus with 256GB. The iPhone App has two ways to connect to the server for video upload. One is through WiFi network in the lab; the other is through 4G mobile network. These two different connections have very different video upload latency. For a typical user video (about one minute long and 30MB in size), the latency of WiFi connection is roughly one second, and the latency of 4G mobile network ranges from a little more than 1 minute to close to 6 minutes. In addition, we found that 4G mobile network was unstable during our experiments. Therefore, the experiments in this section are conducted via WiFi connection.

\setlength{\tabcolsep}{8pt}
\begin{table}[h]
\centering
\caption{The hardware specification for the server.}
    \begin{tabular}{c c}
        \hline
                      & specification\\
        \hline
        CPU           &  AMD Ryzen™ 5 3600X\\
        RAM           &  DDR4-3200 16GB * 2\\
        Graphics Card &  NVIDIA RTX 2060 SUPER\\
        OS            &  Ubuntu 18.04.5 LTS\\
        \hline
    \end{tabular}
\label{tab:server_spec}
\end{table}

\subsection{Performance Metrics}
Before discussing the performance metrics, we first introduce the confusion matrix. The confusion matrix is a popular measure used to handle binary classification problems, such as the pose similarly problem in our system. Table~\ref{tab:c_matrix} shows the confusion matrix, which contains true positive, true negative, false positive, and false negative.

\setlength{\tabcolsep}{5pt}
\begin{table}[h]
\centering
\caption{The confusion matrix.}
    \begin{tabular}{c|c c}
        \hline
                        & Actual: Yes       & Actual: No\\
        \hline
        Predicted: Yes  & True Positive     & False Positive\\
        Predicted: No   & False Negative    & True Negative\\
        \hline
    \end{tabular}
\label{tab:c_matrix}
\end{table}

Based on the confusion matrix, \textbf{Precision}, \textbf{Recall}, and \textbf{F1-Score} are then calculated and used as our performance metrics.

For imbalanced classification problems, Precision and Recall are known to be more appropriate measures. However, high Precision and high Recall are not likely to occur at the same time. If one of them is increased, the other one will be reduced. This situation is called the Precision/Recall tradeoff. To find the right balance between Precision and Recall, F1-Score is used to evaluate our system. In addition, to evaluate the computation of the number of repetitions for an action, two types of accuracy, \textbf{Hard Accuracy} and \textbf{Soft Accuracy}, are defined as follows.

\begin{itemize}
    \item \textit{Hard Accuracy (HA):} Hard Accuracy is defined as the percentage of the number of videos ($CV$), whose number of repetitions is correctly detected, to the number of all videos ($AV$), as shown in Equation~\ref{equ:hard_ac}. For example, given a total of five videos, each having the same action being repeated 10 times. If three of these videos are detected by our system to have 10 number of repetitions and the other two are detected to have more or fewer than 10 repetitions, the $HA$ will be $\frac{3}{5} = 0.6$.

    \begin{equation}
        \label{equ:hard_ac}
        HA = \frac{CV}{AV}
    \end{equation}

    \item \textit{Soft Accuracy (SA):} Soft Accuracy is defined as the percentage of the number of videos (${C_t}V$), whose number of repetitions is detected either correctly or 1 more than the true number of repetitions (i.e., the tolerance is $+$ 1 repetition), to the number of all videos ($AV$), as shown in Equation~\ref{equ:soft_ac}. Because of the definitions of Hard Accuracy and Soft Accuracy, for a given set of videos, we always have $Hard~Accuracy \leq Soft~Accuracy$.

    \begin{equation}
        \label{equ:soft_ac}
        SA = \frac{{C_t}V}{AV}
    \end{equation}
    
\end{itemize}

\subsection{Experimental Results}
In this section, the experimental results of similarity calculation and the number of repetitions will be presented in detail as follows.

\subsubsection{Similarity Calculation}
As mentioned in Section~\ref{sec:dp}, the displacement threshold \(T\) is used to determine if a displacement is large. In our experiments, different \(T\) values are used to calculate the similarity score and evaluate the performance of similarity calculation. Table~\ref{tab:thr_exp} shows the performance of similarity calculation under different \(T\) values.

\begin{figure}[H]
    \begin{center} 
        \includegraphics[width=2in,height=1.2in]{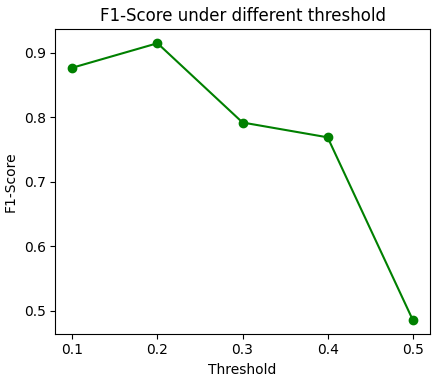}
        \caption{The F1-Score under different threshold.} \label{fig:f1}
    \end{center}
\end{figure}

\setlength{\tabcolsep}{7pt}
\begin{table}[H]
\centering
\caption{The performance of similarity calculation under different threshold.}
    \begin{tabular}{c|c c c}
        \toprule
        \textbf{Threshold}  & \textbf{Precision} & \textbf{Recall} & \textbf{F1-Score}\\
         \hline
        $T = 0.1$ & 0.925 & 0.833 & 0.877\\
        $T = 0.2$ & 0.931 & 0.900 & 0.915\\
        $T = 0.3$ & 0.913 & 0.700 & 0.792\\
        $T = 0.4$ & 0.909 & 0.666 & 0.769\\
        $T = 0.5$ & 0.818 & 0.346 & 0.486\\
        \bottomrule
    \end{tabular}
\label{tab:thr_exp}
\end{table}

Figure~\ref{fig:f1} shows the F1-Score of the similarity calculation under different \(T\) values. As can be seen in the figure, when \(T = 0.2\), the similarity calculation has the highest F1-Score. In addition to F1-Score, Table~\ref{tab:thr_exp} provides the Precision and Recall under different \(T\) values. It can be found that \(T = 0.2\) also gives the best Precision and Recall. This is because the larger the \(T\) value, the smaller the number of coordinates to be revised. On the contrary, when \(T = 0.1\), the number of coordinates that need to be revised may be greater than the number of coordinates revised by other thresholds. In this case, the coordinates will be oversmoothed, resulting in a decrease in F1-Score. Hence, in the following experiments, the value of \(T\) is set to be 0.2.

\subsubsection{The Number of Repetitions}
To evaluate the accuracy of the number of repetitions, several different actions were performed and recorded in the experiments. The collected videos of different actions are listed in Table~\ref{tab:accuracy_int}. As can be seen in the table, these actions are classified into large-range actions and small-range actions. We divide them into two types because it is easier to detect the displacement of the keypoints in a large-range action than a small-range one. Therefore, we would like to examine specifically how well our system performs for small range actions. Note that each video of an action listed in Table~\ref{tab:accuracy_int} may have a different number of repetitions.

Table~\ref{tab:accuracy_exp} presents Hard Accuracy and Soft Accuracy of different actions. As we can see in the table, even though the majority of Hard Accuracy of different actions are only close to 80\%, Soft Accuracy of all actions are above 90\%. In addition, there is no obvious difference in terms of accuracy between large-range actions and small-range actions. In fact, Rotate Neck, one of the actions with the smallest range of movement, is the only action with 100\% Soft Accuracy.

\newcommand{\tabincell}[2]{\begin{tabular}{@{}#1@{}}#2\end{tabular}}  
\setlength{\tabcolsep}{2pt}
\begin{table}[H]
\centering
\caption{Different actions in a large-range and a small-range.}
    \begin{tabular}{c c c c c}
        \toprule
        \tabincell{c}{\textbf{Range of}\\\textbf{action}}
         & \textbf{Action}& $\mathbf{AV}$& $\mathbf{CV}$ & $\mathbf{{C_t}V}$ \\
        \hline
        \multirow{3}{*}{\textbf{Large}} &
        Stand up and squat down & 25 & 19 & 24\\
        &Raise hands & 25 & 21 & 23\\
        &Lift up one foot & 25 & 22 & 23\\
        \hline
        \multirow{3}{*}{\textbf{Small}} &
        Rotate neck & 25 & 23 & 25\\
        &Rotate waist & 25 & 21 & 23\\
        &Shrug shoulders & 25 & 20 & 23\\
        \bottomrule
    \end{tabular}
\label{tab:accuracy_int}
\end{table}

\setlength{\tabcolsep}{3pt}
\begin{table}[H]
\centering
\caption{The accuracy of the actions.}
    \begin{tabular}{c c c}
        \toprule
        \textbf{Action}  & \textbf{Hard Accuracy} &\textbf{Soft Accuracy}\\
        \hline
        Stand up and squat down & 0.76 & 0.96\\
        Raise hands & 0.84 & 0.92\\
        Lift up one foot & 0.88 & 0.92\\
        \hline
        Rotate neck & 0.92 & 1.00\\
        Rotate waist & 0.84 & 0.92\\
        Shrug shoulders & 0.80 & 0.92\\
        \bottomrule
    \end{tabular}
\label{tab:accuracy_exp}
\end{table}

\section{Conclusion}
\label{sec:Conclusion}
In this paper, we propose an HRNet-based rehabilitation monitoring system. The purpose is to build a monitoring mechanism between the therapist and the patient. Our proposed system consists of the client app, the web server, and HRNet-based action monitoring module. The client app is used by the therapist and patient to record the action and upload the video to the server. The web server is used to store videos, add rehabilitation actions, notify patients of their scheduled rehabilitation actions, and calculate the completion rate. The HRNet-based action monitoring module is used to extract pose features from the patient's uploaded video, detect incorrect coordinates reported by the HRNet model, calculate the similarity score using KL-divergence, and derive the number of repetitions of different actions. The results of extensive experiments indicate that the F1-Score of the similarity calculation is as high as 0.9 and the soft accuracy of the number of repetitions for any action is higher than 90\%.

\bibliographystyle{named}
\bibliography{Reference}

\end{document}